\pdfoutput=1

\documentclass[11pt]{article}

\usepackage{naacl2021}

\usepackage{times}
\usepackage{latexsym}

\usepackage[T1]{fontenc}

\usepackage[utf8]{inputenc}

\usepackage{microtype}

\usepackage{pgf}

\title{Textgraphs-15 Shared Task System Description : Multi-Hop Inference Explanation Regeneration by Matching Expert Ratings}

\author{Vivek Kalyan 
  \thanks{\hspace{2mm} Work done in conjunction with Red Dragon AI}\\ 
  Singapore \\
  {\tt hello@vivekkalyan.com} \\
  \And
  Sam Witteveen \\
  Red Dragon AI \\
  Singapore \\
  {\tt sam@reddragon.ai} \\\And
  Martin Andrews
  \thanks{\hspace{2mm} Corresponding author}\\ 
  Red Dragon AI  \\
  Singapore \\
  {\tt martin@reddragon.ai} \\}

\begin{document}
\maketitle
\begin{abstract}
Creating explanations for answers to science questions is a challenging task that requires multi-hop inference over a large set of fact sentences.  
This year, to refocus the Textgraphs Shared Task on the problem of gathering relevant statements (rather than solely finding a single `correct path'), the WorldTree dataset was augmented with expert ratings of `relevance' of statements to each overall explanation.  
Our system, which achieved second place on the Shared Task leaderboard, combines initial statement retrieval; language models trained to predict the relevance scores; and ensembling of a number of the resulting rankings.
Our code implementation is made available at
\url{https://github.com/mdda/worldtree_corpus/tree/textgraphs_2021}
\end{abstract}

\section{Introduction}

Complex question answering often requires reasoning over many evidence documents, which is known as multi-hop inference.
Existing datasets such as Wikihop \cite{welbl-etal-2018-constructing}, OpenBookQA \cite{OpenBookQA2018}, QASC \cite{khot2020qasc}, are limited due to artificial questions and short aggregation, requiring less than 3 facts.
In comparison, the TextGraphs Shared Task \cite{Jansen:20} makes use of 
WorldTree V2 \cite{xie-etal-2020-worldtree} which a large dataset of over 5,000 questions and answers, as well as detailed explanations that link them.
The `gold' explanation paths require combining an average of 6 and up to 16 facts in order to generate an full explanation for complex science questions.  

The WorldTree dataset was recently supplemented with 
approximately 250,000 expert-annotated relevancy ratings 
for facts that were highly ranked by models in previous Shared Task iterations,
based on the same consistent set of question and answers.  






In previous years, the emphasis of the Shared Task has been on creating `connected explanations' as completely as possible, which is difficult because of the large branching factor along an explanation path, in conjunction with semantic drift \cite{fried-etal-2015-higher}.  
In contrast, 
the scoring function for the 2021 Shared Task required participants to rank explanation statements according to their relevance to explaining the science situation, rather than whether they are in the single gold explanation path.
Specifically, 
participants were required to provide ordered lists of explanation statements for each question, 
and the Normalized Discounted Cumulative Gain measure 
(`NDCG' - \citeauthor{burges-etal-2005}, \citeyear{burges-etal-2005}) 
was used as a scoring function.  

\begin{figure*}[t!]
    \centering
    \input{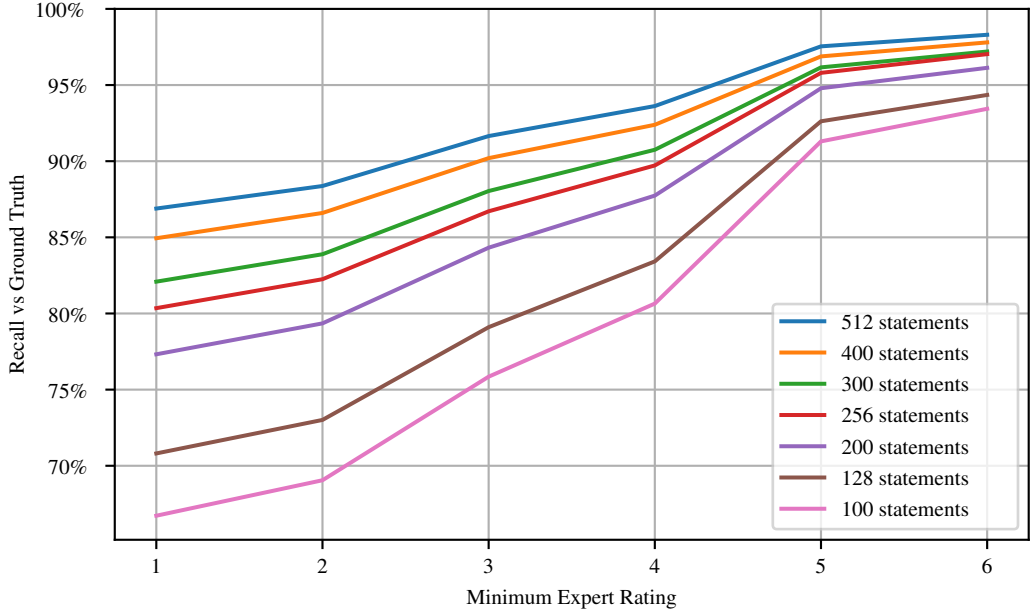}
    \caption{Recall by Rating for different numbers of statements retrieved by I-BM25 stage}
    \label{graph:recall}
\end{figure*}

The main contributions of this work are:
\begin{enumerate}
    \item We show that conventional information retrieval-based methods are still a strong baseline and use a hyperparameter-optimised version of I-BM25, an iterative retrieval method that improves inference speed and recall by emulating multi-hop retrieval.
    \item We propose a simple BERT-based architecture that predicts the expert rating of each explanation statement in the context of the current question (and correct answer).
    \item We ensemble language model rankings in order to increase our leaderboard score.
\end{enumerate}

\section{Models}

Neural information retrieval models such as DPR \cite{karpukhin2020dense}, RAG \cite{lewis2020retrieval}, and ColBERT \cite{khattab2020colbert} that assume query-document independence use a language model to generate sentence representations for the query and document separately.
The advantage of this late-interaction approach is efficient inference as the sentence representations can be computed beforehand and optimized lookup methods such as FAISS \cite{JDH17} exist for this purpose.
However, the late-interaction compromises on deeper semantic understanding possible with language models.
Early-interaction approaches such as TFR-BERT \cite{han2020learning} instead concatenate the query and document before generating a unified sentence representation. 
This approach is more computationally expensive but is attractive for re-ranking over a limited number of documents.
%
%
To reduce this computational burden, we have a front-end to our system that retrieves a limited number of facts for later re-ranking by language models.  

Overall, our final system comprised 3 distinct stages, each of which were tailored to the Shared Task : Initial retrieval, Language Models and final ensembling.


\subsection{Iterative BM25 Retrieval}

Chia et al \shortcite{chia-etal-2019-red} and Chia et al \shortcite{chia-etal-2020-red} showed that conventional information retrieval methods can be a strong baseline when modified to suit the multi-hop inference objective.

We adapted the iterative retrieval method (denoted `I-BM25') from 
Chia et al \shortcite{chia-etal-2020-red} 
that was shown to perform inference quickly and reduce the impact of semantic drift, 
resulting in a strong retrieval method for subsequent re-ranking.
For preprocessing, we use spaCy \cite{spacy2} for tokenization, lemmatization and stopword removal.
%
\filbreak
The I-BM25 algorithm is as follows:
\begin{enumerate}
    \item Sparse document vectors are pre-computed for all questions and explanation candidates.
    \item For each question, the closest $n$ explanation candidates by cosine proximity are selected and their vectors are aggregated by a $max$ operation.
    The aggregated vector is down-scaled and used to update the query vector through a $max$ operation.
    \item The previous step is repeated for increasing values of $n$ until there are no candidate explanations remaining.
\end{enumerate}


Included within the algorithm above are a number of hyperparameters 
(such as the rate of increase of $n$, and parameters of the BM25 search framework)
which were previously optimised for their performance on the 2020 TextGraphs Shared Task.  
These were re-optimised for the 2021 version, 
with the goal of maximising the average recall over each category of expert score, 
for a given number of retrieved explanation statements.
Using Figure \ref{graph:recall}, 
the number of retrieved statements was chosen to be 200 in the interests of balancing
overall recall (93.78\% of statements with scores higher than zero) 
with the later processing cost imposed by the length of the list of candidate statements.

\begin{table*}[ht]
\centering
\begin{tabular}{|l  | r r |} 
 \hline
 Model & Dev NDCG & Test NDCG \\
 \hline 
 Baseline TF-IDF & 0.5130 & 0.5010 \\
 I-BM25-base & 0.6669 & n/a \\ 
 I-BM25 & 0.6785 & 0.6583 \\ 
 I-BM25 + BERT & 0.7679 & 0.7580 \\
 I-BM25 + BERT ensemble & 0.7801 & 0.7675 \\
 I-BM25 + BERT + SciBERT ensemble & 0.7836 & 0.7705 \\
\hline
\end{tabular}
\caption{NDCG score comparison as evaluated locally and on the leaderboard}
\label{table:1}
\end{table*}

\subsection{Language Models for Rating Classification/Regression}
Pre-trained versions of BERT \cite{devlin2019bert} are widely 
adapted and fine-tuned for many downstream NLP tasks.  
For the Shared Task, we fine-tuned this language model 
to predict the Expert Rating from text sequences, 
where each sequence is a question (including the correct answer) and explanation pair separated by the [SEP] token, 
and the prediction task is a regression against the gold Expert Rating (using a Mean Square Error loss minimisation objective).


During inference, we use the 200 explanations returned by the earlier 
I-BM25 phase for each question, fed into BERT as a question and explanation pair.  
We then used the (floating point) score output by the trained BERT 
as a sortable value by which to rank the explanations in terms of relevancy.

\subsection{Ensembling of Rankings}

In the later stages of the competition, 
we decided to employ an ensemble of different models - 
%
4 BERT models (each fine-tuned with a different seed) 
and a similarly fine-tuned model based on a pretrained SciBERT \cite{Beltagy2019SciBERT}.

We ensembled the ranked output of each model together by simply linearly combining 
each rank into an aggregate.\footnote{This method was simplified 
since each of the re-rankings was sourced from the same I-BM25 output list}  
More sophisticated combinations were considered, but these suffered from overfitting on the Dev set.


\section{Experiments}

Our system comprised three stages, 
and we present results of the experiments used to validate our choices at each stage,
with the overall results being compiled in Table \ref{table:1}.

\subsection{Retrieval}
As an initial step, we focused on ensuring our retrieval model found as many relevant explanations as possible in its output list (regardless of the order), while keeping the list as short as possible.
So as to measure this, we computed an ``Oracle NDCG'' score,
the score the retrieval model would have received if it had access to an oracle and thus could return the perfect rank ordering.

\begin{table}[h!]
\centering
\begin{tabular}{|l  | r |} 
 \hline
 Retrieval Model & Oracle NDCG \\
 \hline
 TF-IDF     & 0.7547 \\   
 I-BM25-base     & 0.8941 \\ 
 I-BM25     & 0.9378 \\ 
\hline
\end{tabular}
\caption{Oracle NDCG score on WorldTree V2 dataset}
\label{table:2}
\end{table}

In addition to measuring the performance of the initial retrieval stage, 
the Oracle NDCG score also gave us the ceiling for performance of 
our second stage models.
%

\subsection{Language Models}


\begin{table}[h!]
\centering
\begin{tabular}{|l  | r |} 
 \hline
 Language Model & Dev NDCG \\
 \hline
 DistilBERT & 0.7353 \\   
 BERT       & 0.7679 \\ 
 SciBERT    & 0.7541 \\ 
\hline
\end{tabular}
\caption{Language model comparison}
\label{table:3}
\end{table}

While we initially tried DistilBERT \cite{sanh2020distilbert} 
- a lean version of BERT with fewer parameters - 
we found that BERT outperformed DistilBERT by a significant enough margin
to suggest that the efficiency of DistilBERT was not a net win.

We also attempted to fine-tune RoBERTa \cite{liu2019roberta} on the regression task, 
but were unable to achieve satisfactory results quickly enough 
to incorporate it into our ensembling regime. 


\subsection{Ensembling}

While SciBERT performed slightly worse than other models on an individual basis, 
ensembling it with regular BERT models resulted in a much higher score -
which suggests that its representations are well differentiated by its pretraining regime.

\section{Negative Results}

\subsection{Two-stage representation}
In addition to the straight regression models used in our final submissions, 
we also investigated an architecture that modelled the explanation ratings for each question/answer via a two-stage process.  

The first stage was a binary indicator of whether the explanation was relevant or not 
($1$ if it had a higher-than-zero rating, $0$ if zero-rated or missing).  
The second stage (used during inference if the first stage signalled `relevant'), 
was modelled as a distribution over the possible scores.  
The intuition being that some statements are `broad, powerful concepts' (likely to score highly if relevant) whereas others are `tiny lexical adjustments' (likely to be low-scoring if considered relevant).

Despite the intuitive appeal of modelling the statement rating process in this way, 
and the apparently reasonable distributions learned, 
this architecture did not lead to higher scores overall - though that may be due to other factors 
(such as running out of time to finesse the training and/or inference process).

\subsection{Negative Sampling}

While examining the types of prediction errors our initial models were making during inference, 
we noticed that quite a number of the incorrectly chosen explanations (from the I-BM25 stage)
were lexically close to highly-rated explanation statements.
%
%
This showed that there was a mismatch between frequency of 
zero-rated Expert Ratings in the Train set, and what would be experienced during inference.
Therefore, we hypothesised that adding more negative samples 
would help the model discern between these similar explanations.


Two methods were tried : (i) Randomly sampling from the explanations database; 
and (ii) Using the retrieval model to propose other close negatives during training.
Unfortunately, neither resulted in any significant improvement in scores.

\section{Discussion}

In previous versions of the Textgraphs Shared Task, 
the goal was essentially to obtain the single `gold explanation' that perfectly matched an expertly
crafted graph of explanation statements, 
with the scoring being based on a ranking metric
that rewarded participants for finding these gold explanation statements.
%
This task was challenging due to the semantic drift issue previously mentioned,
and the sensitivity of the scoring to choosing the same explanation path as the original 
annotators.\footnote{In terms of extra data, 
supposing that a Worldtree Explanation Corpus continues 
to be the basis of the Textgraphs Shared Task in the future, 
it would be very helpful to have the structured information
that resulted in the output of the Worldtree Explanation Corpus v2.1 Desk Reference, 
since that would allow a cleaner interpretation of the structured table data - 
without participants having to each independently reinvent the wheel}
%
Paradoxically, 
instead of tackling the problem with logic-oriented graph planning methods,
the dominant techniques tended to rely on large language models 
which could maximise the ranking scores without `understanding the bigger picture'.


The change of scoring metric in this current Shared Task, 
to incorporate all statements that are relevant to the question and answer, 
appears to target the capturing of `bigger picture' ideas.
However, 
this seems to have once again promoted the use of large language models,
since they provide a system component that can bring the most `common sense' 
into the multi-step reasoning domain, 
without getting tangled in the logical weeds that go into producing the gold explanations.

While the addition of the expert ratings on the explanation statements 
is undoubtedly positive for the Shared Task dataset, 
it is not clear to what extent it helps address the multi-hop nature of the challenge - 
on which significant progress had already been made
(and will hopefully continue based on other promising directions have been 
identified by previous iterations of the Shared Task.)

\section{Conclusion}

Our Shared Task submissions showed that ensembles of language models 
trained on a regression basis to predict Expert Ratings 
obtain highly competitive results.

We look forward to achieving further progress on the multi-hop reasoning task in the future.


\filbreak

\bibliography{anthology,custom}

\begin{thebibliography}{19}
\expandafter\ifx\csname natexlab\endcsname\relax\def\natexlab#1{#1}\fi

\bibitem[{Beltagy et~al.(2019)Beltagy, Lo, and Cohan}]{Beltagy2019SciBERT}
Iz~Beltagy, Kyle Lo, and Arman Cohan. 2019.
\newblock \href {http://arxiv.org/abs/1903.10676} {Sci{BERT}: Pretrained
  language model for scientific text}.
\newblock \emph{arXiv preprint arXiv:1903.10676}.

\bibitem[{Burges et~al.(2005)Burges, Shaked, Renshaw, Lazier, Deeds, Hamilton,
  and Hullender}]{burges-etal-2005}
Chris Burges, Tal Shaked, Erin Renshaw, Ari Lazier, Matt Deeds, Nicole
  Hamilton, and Greg Hullender. 2005.
\newblock \href {https://doi.org/10.1145/1102351.1102363} {Learning to rank
  using gradient descent}.
\newblock In \emph{Proceedings of the 22nd International Conference on Machine
  Learning}, ICML '05, page 89–96, New York, NY, USA. Association for
  Computing Machinery.

\bibitem[{Chia et~al.(2019)Chia, Witteveen, and Andrews}]{chia-etal-2019-red}
Yew~Ken Chia, Sam Witteveen, and Martin Andrews. 2019.
\newblock \href {https://doi.org/10.18653/v1/D19-5311} {Red {D}ragon {AI} at
  {T}ext{G}raphs 2019 shared task: Language model assisted explanation
  generation}.
\newblock In \emph{Proceedings of the Thirteenth Workshop on Graph-Based
  Methods for Natural Language Processing (TextGraphs-13)}, pages 85--89, Hong
  Kong. Association for Computational Linguistics.

\bibitem[{Chia et~al.(2020)Chia, Witteveen, and Andrews}]{chia-etal-2020-red}
Yew~Ken Chia, Sam Witteveen, and Martin Andrews. 2020.
\newblock \href {https://www.aclweb.org/anthology/2020.textgraphs-1.14} {Red
  {D}ragon {AI} at {T}ext{G}raphs 2020 shared task : {LIT} : {LSTM}-interleaved
  transformer for multi-hop explanation ranking}.
\newblock In \emph{Proceedings of the Graph-based Methods for Natural Language
  Processing (TextGraphs)}, pages 115--120, Barcelona, Spain (Online).
  Association for Computational Linguistics.

\bibitem[{Devlin et~al.(2019)Devlin, Chang, Lee, and
  Toutanova}]{devlin2019bert}
Jacob Devlin, Ming-Wei Chang, Kenton Lee, and Kristina Toutanova. 2019.
\newblock \href {http://arxiv.org/abs/1810.04805} {{BERT}: Pre-training of deep
  bidirectional transformers for language understanding}.
\newblock \emph{arXiv preprint arXiv:1810.04805}.

\bibitem[{Fried et~al.(2015)Fried, Jansen, Hahn-Powell, Surdeanu, and
  Clark}]{fried-etal-2015-higher}
Daniel Fried, Peter Jansen, Gustave Hahn-Powell, Mihai Surdeanu, and Peter
  Clark. 2015.
\newblock \href {https://doi.org/10.1162/tacl_a_00133} {Higher-order lexical
  semantic models for non-factoid answer reranking}.
\newblock \emph{Transactions of the Association for Computational Linguistics},
  3:197--210.

\bibitem[{Han et~al.(2020)Han, Wang, Bendersky, and Najork}]{han2020learning}
Shuguang Han, Xuanhui Wang, Mike Bendersky, and Marc Najork. 2020.
\newblock \href {http://arxiv.org/abs/2004.08476} {Learning-to-rank with {BERT}
  in {TF-R}anking}.
\newblock \emph{arXiv preprint arXiv:2004.08476}.

\bibitem[{Honnibal and Montani(2017)}]{spacy2}
Matthew Honnibal and Ines Montani. 2017.
\newblock {spaCy 2}: Natural language understanding with {B}loom embeddings,
  convolutional neural networks and incremental parsing.
\newblock To appear.

\bibitem[{Jansen and Ustalov(2020)}]{Jansen:20}
Peter Jansen and Dmitry Ustalov. 2020.
\newblock \href {https://www.aclweb.org/anthology/2020.textgraphs-1.10}
  {{TextGraphs~2020 Shared Task on Multi-Hop Inference for Explanation
  Regeneration}}.
\newblock In \emph{Proceedings of the Graph-based Methods for Natural Language
  Processing (TextGraphs)}, pages 85--97, Barcelona, Spain (Online).
  Association for Computational Linguistics.

\bibitem[{Johnson et~al.(2017)Johnson, Douze, and J{\'e}gou}]{JDH17}
Jeff Johnson, Matthijs Douze, and Herv{\'e} J{\'e}gou. 2017.
\newblock \href {http://arxiv.org/abs/1702.08734} {Billion-scale similarity
  search with {GPU}s}.
\newblock \emph{arXiv preprint arXiv:1702.08734}.

\bibitem[{Karpukhin et~al.(2020)Karpukhin, O{\u{g}}uz, Min, Wu, Edunov, Chen,
  and Yih}]{karpukhin2020dense}
Vladimir Karpukhin, Barlas O{\u{g}}uz, Sewon Min, Ledell Wu, Sergey Edunov,
  Danqi Chen, and Wen-tau Yih. 2020.
\newblock \href {http://arxiv.org/abs/2004.04906} {Dense passage retrieval for
  open-domain question answering}.
\newblock \emph{arXiv preprint arXiv:2004.04906}.

\bibitem[{Khattab and Zaharia(2020)}]{khattab2020colbert}
Omar Khattab and Matei Zaharia. 2020.
\newblock \href {http://arxiv.org/abs/2004.12832} {Col{BERT}: Efficient and
  effective passage search via contextualized late interaction over {BERT}}.
\newblock \emph{arXiv preprint arXiv:2004.12832}.

\bibitem[{Khot et~al.(2020)Khot, Clark, Guerquin, Jansen, and
  Sabharwal}]{khot2020qasc}
Tushar Khot, Peter Clark, Michal Guerquin, Peter Jansen, and Ashish Sabharwal.
  2020.
\newblock {QASC}: A dataset for question answering via sentence composition.
\newblock In \emph{AAAI}, pages 8082--8090.

\bibitem[{Lewis et~al.(2020)Lewis, Perez, Piktus, Petroni, Karpukhin, Goyal,
  K{\"u}ttler, Lewis, Yih, Rockt{\"a}schel et~al.}]{lewis2020retrieval}
Patrick Lewis, Ethan Perez, Aleksandara Piktus, Fabio Petroni, Vladimir
  Karpukhin, Naman Goyal, Heinrich K{\"u}ttler, Mike Lewis, Wen-tau Yih, Tim
  Rockt{\"a}schel, et~al. 2020.
\newblock \href {http://arxiv.org/abs/2005.11401} {Retrieval-augmented
  generation for knowledge-intensive {NLP} tasks}.
\newblock \emph{arXiv preprint arXiv:2005.11401}.

\bibitem[{Liu et~al.(2019)Liu, Ott, Goyal, Du, Joshi, Chen, Levy, Lewis,
  Zettlemoyer, and Stoyanov}]{liu2019roberta}
Yinhan Liu, Myle Ott, Naman Goyal, Jingfei Du, Mandar Joshi, Danqi Chen, Omer
  Levy, Mike Lewis, Luke Zettlemoyer, and Veselin Stoyanov. 2019.
\newblock \href {http://arxiv.org/abs/1907.11692} {Ro{BERT}a: A robustly
  optimized {BERT} pretraining approach}.
\newblock \emph{arXiv preprint arXiv:1907.11692}.

\bibitem[{Mihaylov et~al.(2018)Mihaylov, Clark, Khot, and
  Sabharwal}]{OpenBookQA2018}
Todor Mihaylov, Peter Clark, Tushar Khot, and Ashish Sabharwal. 2018.
\newblock \href {http://arxiv.org/abs/1809.02789} {Can a suit of armor conduct
  electricity? {A} new dataset for open book question answering}.
\newblock \emph{arXiv preprint arXiv:1809.02789}.

\bibitem[{Sanh et~al.(2020)Sanh, Debut, Chaumond, and
  Wolf}]{sanh2020distilbert}
Victor Sanh, Lysandre Debut, Julien Chaumond, and Thomas Wolf. 2020.
\newblock \href {http://arxiv.org/abs/1910.01108} {Distil{BERT}, a distilled
  version of {BERT}: smaller, faster, cheaper and lighter}.
\newblock \emph{arXiv preprint arXiv:1910.01108}.

\bibitem[{Welbl et~al.(2018)Welbl, Stenetorp, and
  Riedel}]{welbl-etal-2018-constructing}
Johannes Welbl, Pontus Stenetorp, and Sebastian Riedel. 2018.
\newblock \href {https://doi.org/10.1162/tacl_a_00021} {Constructing datasets
  for multi-hop reading comprehension across documents}.
\newblock \emph{Transactions of the Association for Computational Linguistics},
  6:287--302.

\bibitem[{Xie et~al.(2020)Xie, Thiem, Martin, Wainwright, Marmorstein, and
  Jansen}]{xie-etal-2020-worldtree}
Zhengnan Xie, Sebastian Thiem, Jaycie Martin, Elizabeth Wainwright, Steven
  Marmorstein, and Peter Jansen. 2020.
\newblock \href {https://www.aclweb.org/anthology/2020.lrec-1.671}
  {{W}orld{T}ree v2: A corpus of science-domain structured explanations and
  inference patterns supporting multi-hop inference}.
\newblock In \emph{Proceedings of the 12th Language Resources and Evaluation
  Conference}, pages 5456--5473, Marseille, France. European Language Resources
  Association.

\end{thebibliography}
\bibliographystyle{acl_natbib}




\end{document}